\pdfoutput=1

\documentclass[11pt]{article}
\usepackage[table]{xcolor}

\usepackage[final]{acl}

\usepackage{times}
\usepackage{latexsym}

\usepackage[T1]{fontenc}

\usepackage[utf8]{inputenc}

\usepackage{microtype}

\usepackage{inconsolata}

\usepackage{microtype}

\usepackage{inconsolata}

\usepackage{appendix}
\usepackage{makecell}
\usepackage{booktabs}
\usepackage{arydshln}
\usepackage{amsmath}
\usepackage{amssymb}
\usepackage{graphicx}
\usepackage{fontawesome5}
\usepackage{multirow}
\usepackage{cleveref}
\usepackage{enumitem}
\usepackage{amsfonts}
\usepackage{subcaption}
\usepackage{algorithm}
\usepackage{algorithmicx}
\usepackage[noend]{algpseudocode}
\usepackage{soul}
\usepackage{diagbox}
\usepackage{enumitem}
\usepackage{tcolorbox}
\usepackage{caption}
\usepackage{subcaption}

\crefname{section}{\S}{\S\S}
\Crefname{section}{\S}{\S\S}
\crefname{appendix}{Appendix}{Appendices}
\Crefname{appendix}{Appendix}{Appendices}
\crefformat{equation}{Eq.~#2#1#3}
\Crefformat{equation}{Equation~#2#1#3}

\definecolor{hzw}{RGB}{223, 97, 76}
\sethlcolor{Melon}

\definecolor{xing}{rgb}{1.0, 0.03, 0.0}

\definecolor{nred}{RGB}{196, 38, 11}
\definecolor{nblue}{RGB}{41, 52, 190}
\definecolor{lightgreen}{RGB}{230, 238, 227}
\definecolor{ngreen}{RGB}{18, 141, 21}
\definecolor{sblue}{RGB}{34, 48, 78}

\usepackage{CJKutf8}

\newcommand{\hollowSquare}{\scalebox{0.9}{\textcolor{sblue}{\faSquare[regular]}}}
\newcommand{\solidSquare}{\scalebox{0.9}{\textcolor{sblue}{\faSquare}}}
\newcommand{\hollowStar}{\scalebox{0.9}{\textcolor{sblue}{\faStar[regular]}}}
\newcommand{\solidStar}{\scalebox{0.9}{\textcolor{sblue}{\faStar}}}
\newcommand{\hollowHeart}{\scalebox{0.9}{\textcolor{sblue}{\faHeart[regular]}}}
\newcommand{\solidHeart}{\scalebox{0.9}{\textcolor{sblue}{\faHeart}}}

\makeatletter
\DeclareRobustCommand{\iscircle}{\mathord{\mathpalette\is@circle\relax}}
\newcommand\is@circle[2]{%
  \begingroup
  \sbox\z@{\raisebox{\depth}{$\m@th#1\bigcirc$}}%
  \sbox\tw@{$#1\square$}%
  \resizebox{!}{\ht\tw@}{\usebox{\z@}}%
  \endgroup
}
\makeatother

\title{Can Watermarks Survive Translation?\\On the Cross-lingual Consistency of Text Watermark\\for Large Language Models}

\author{%
    Zhiwei He$^1$\thanks{Equal Contribution. Work done during Zhiwei's internship at Tencent AI Lab.}\quad Binglin Zhou$^1$$^*$\quad Hongkun Hao$^1$\quad Aiwei Liu$^3$\\
    \bf Xing Wang$^2$$^\dagger$\quad Zhaopeng Tu$^2$\quad Zhuosheng Zhang$^1$\quad Rui Wang$^1$\thanks{Rui Wang and Xing Wang are co‐corresponding authors.}\\
    $^1$Shanghai Jiao Tong University\ \ \ $^2$Tencent AI Lab\ \ \ $^3$Tsinghua University\\
    $^1$\texttt{\small{\{zwhe.cs,zhoubinglin,zhangzs,wangrui12\}}@sjtu.edu.cn} \\
    $^2$\texttt{\small{\{brightxwang,zptu\}}@tencent.com}\quad$^3$\texttt{\small{liuaw20@mails.tsinghua.edu.cn}} \\
    \texttt{\small{{\faGithub}[Official]:\url{https://github.com/zwhe99/X-SIR}\quad{\faGithub}[Toolkit]:\url{https://github.com/THU-BPM/MarkLLM}}} \\
}

\begin{document}
\maketitle
\begin{abstract}
Text watermarking technology aims to tag and identify content produced by large language models (LLMs) to prevent misuse.
In this study, we introduce the concept of ``{\em cross-lingual consistency}'' in text watermarking, which assesses the ability of text watermarks to maintain their effectiveness after being translated into other languages.
Preliminary empirical results from two LLMs and three watermarking methods reveal that current text watermarking technologies lack consistency when texts are translated into various languages.
Based on this observation, we propose a Cross-lingual Watermark Removal Attack (CWRA) to bypass watermarking by first obtaining a response from an LLM in a pivot language, which is then translated into the target language.
CWRA can effectively remove watermarks, decreasing the AUCs to a random-guessing level without performance loss.
Furthermore, we analyze two key factors that contribute to the cross-lingual consistency in text watermarking and propose X-SIR as a defense method against CWRA.
\end{abstract}
\section{Introduction}

\begin{figure}[t!]
    \centering
    \includegraphics[width=0.82\linewidth]{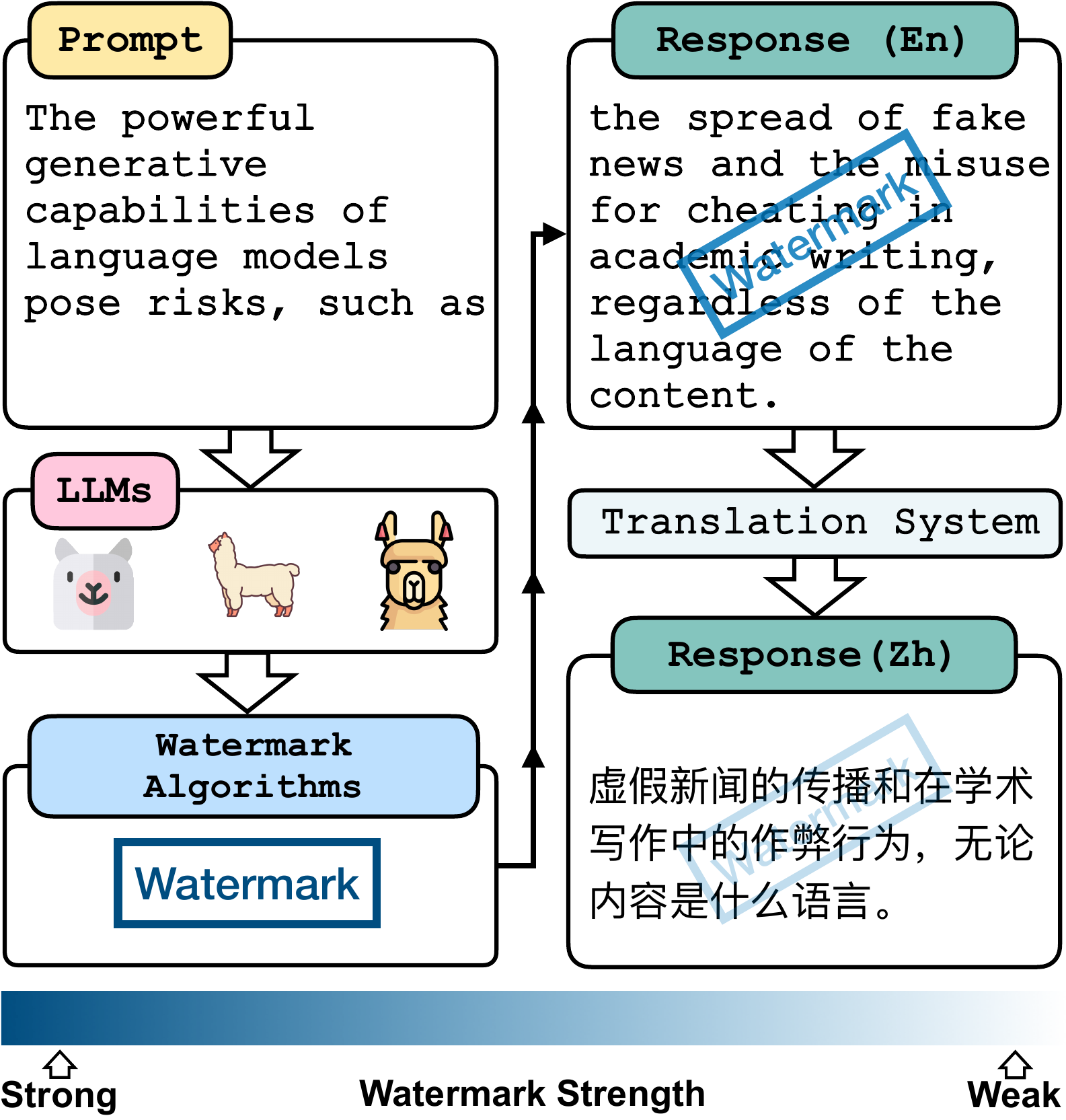}
    \caption{Illustration of watermark dilution in a cross-lingual environment. Best viewed in color.}
    \label{fig:intro}
\end{figure}
Large language models (LLMs) like GPT-4~\cite{openai2023gpt4} have demonstrated remarkable content generation capabilities, producing texts that are hard to distinguish from human-written ones.
This progress has led to concerns regarding the misuse of LLMs, such as the risks of generating misleading information, impersonating individuals, and compromising academic integrity~\cite{chen2023can,ai2024cognition,yuan2024rjudge,xia2024measuring}.
As a countermeasure, text watermarking technology for LLMs has been developed, aiming at tagging and identifying the content produced by LLMs~\cite{pmlr-v202-kirchenbauer23a,liu2023survey}.
Generally, a text watermarking algorithm embeds a message within LLM-generated content that is imperceptible to human readers, but can be detected algorithmically.
By tracking and detecting text watermarks, it becomes possible to mitigate the abuse of LLMs by tracing the origin of texts and ascertaining their authenticity.

The robustness of watermarking algorithms, i.e., the ability to detect watermarked text even after it has been modified, is important.
Recent works have shown strong robustness under text rewriting and copy-paste attacks~\cite{liu2023semantic,yang2023watermarking}.
However, these watermarking techniques have been tested solely within monolingual contexts.
In practical scenarios, watermarked texts might be translated~\cite{10.1162/tacl_a_00642,he2023feedbackmt,he-etal-2022-bridging,jiao-etal-2023-parrot,liang2023encouraging}, raising questions about the efficacy of text watermarks across languages (see \figurename~\ref{fig:intro}).
For example, a malicious user could use a watermarked LLM to produce fake news in English and then translate it into Chinese.
Obviously, the deceptive impact persists regardless of the language, but it is uncertain whether the watermark would still be detectable after such a translation.
To explore this question, we introduce the concept of \textit{cross-lingual consistency} in text watermarking, aiming to characterize the ability of text watermarks to preserve their strength across languages.
Our preliminary results on 2 LLMs $\times$ 3 watermarks reveal that current text watermarking technologies lack consistency across languages.

In light of this finding, we propose the \textbf{C}ross-lingual \textbf{W}atermark \textbf{R}emoval \textbf{A}ttack (\textbf{CWRA}) to highlight the practical implications arising from deficient cross-lingual consistency.
When performing CWRA, the attacker begins by translating the original language prompt into a pivot language, which is fed to the LLM to generate a response in the pivot language.
Finally, the response is translated back into the original language.
In this way, the attacker obtains the response in the original language and bypasses the watermark with the second translation step.
CWRA outperforms re-writing attacks, such as re-translation and paraphrasing~\cite{liu2023survey}, as it decreases the AUCs to a random-guessing level and achieves the highest text quality.

To resist CWRA, we propose a defense method that improves the cross-lingual consistency of current LLM watermarking.
Our method is based on two critical factors.
The first is the \textbf{cross-lingual semantic clustering of the vocabulary}.
Instead of treating each token in the vocabulary as the smallest unit when ironing watermarks, as done by KGW~\cite{pmlr-v202-kirchenbauer23a}, our method considers a cluster of tokens that share the same semantics across different languages as the smallest unit of processing.
In this way, the post-translated token will still carry the watermark as it would fall in the same cluster as before translation.
The second is \textbf{cross-lingual semantic robust vocabulary partition}.
Inspired by \citet{liu2023semantic}, we ensure that the partition of the vocabulary are similar for semantically similar contexts in different languages.
Despite its limitations, our approach (named X-SIR) substantially elevates the AUCs under the CWRA, paving the way for future research.

Our contributions are summarized as follows:
\begin{itemize}[topsep=0pt, partopsep=0pt,itemsep=0pt,parsep=0pt,leftmargin=10pt]
    \item \textbf{Evaluation} (\cref{sec:cross-lingual-consistency-of-text-watermark}): We reveal the deficiency of current text watermarking technologies in maintaining cross-lingual consistency.

    \item \textbf{Attack} (\cref{sec:cross-lingual-watermark-removal-attack}): Based on this finding, we propose CWRA that successfully bypasses watermarks without degrading the text quality.

    \item \textbf{Defense} (\cref{sec:improving-cross-lingual-consistency}): We identify two key factors for improving cross-lingual consistency and propose X-SIR as a defense method against CWRA.
\end{itemize}
\section{Background}
\subsection{Language Model}
A language model (LM) $M$ has a defined set of tokens known as its vocabulary $\mathcal{V}$.
Given a sequence of tokens $\boldsymbol{x}^{1:n}=(x^1, x^2, \ldots, x^n)$, which we refer to as the \textit{prompt}, the model $M$ computes the conditional probability of the next token over $\mathcal{V}$ as $P_{M}(x^{n+1} | \boldsymbol{x}^{1:n})$.
Therefore, text generation can be achieved through an autoregressive decoding process, where $M$ sequentially predicts one token at a time, forming a \textit{response}.
Such an LM can be parameterized by a neural network, such as Transformer~\cite{NIPS2017_3f5ee243}, which is called neural LM.
Typically, a neural LM computes a vector of logits $\boldsymbol{z}^{n+1}=M(\boldsymbol{x}^{1:n}) \in \mathbb{R}^{|\mathcal{V}|}$ for the next token based on the current sequence $\boldsymbol{x}_{1:n}$ via a neural network.
The probability of the next token is then obtained by applying the softmax function to these logits: $P_{M}(x^{n+1} | \boldsymbol{x}^{1:n}) = \text{softmax}(\boldsymbol{z}^{n+1})$.

\subsection{Watermarking for LMs}
\label{sec:watermarking-for-lms}
In this work, we consider the following watermarking methods.
All of them embed the watermark by modifying logits during text generation and detect the presence of the watermark for any given text.

\textbf{KGW}~\cite{pmlr-v202-kirchenbauer23a} sets the groundwork for LM watermarking.
Ironing a watermark is delineated as the following steps:
\begin{enumerate}[topsep=0pt, partopsep=0pt,itemsep=0pt,parsep=0pt]
    \item[(1)] compute a hash of $\boldsymbol{x}^{1:n}$: $h^{n+1}=H(\boldsymbol{x}^{1:n})$,
    \item[(2)] seed a random number generator with $h^{n+1}$ and randomly partitions $\mathcal{V}$ into two disjoint lists: the \textit{green} list $\mathcal{V}_g$ and the \textit{red} list $\mathcal{V}_r$,
    \item[(3)] adjust the logits $\boldsymbol{z}^{n+1}$ by adding a constant bias $\delta$ ($\delta > 0$) for tokens in the green list:
    \begin{equation}
    \begin{aligned}
        &\forall i \in \{1,2,\dots,|\mathcal{V}|\}, \\
        &\boldsymbol{\tilde{z}}^{n+1}_{i} =
        \begin{cases}
        \boldsymbol{z}^{n+1}_{i} + \delta, & \text{if } v_i \in \mathcal{V}_g, \\
        \boldsymbol{z}^{n+1}_{i}, & \text{if } v_i \in \mathcal{V}_r.
        \end{cases}
    \end{aligned}
    \end{equation}
\end{enumerate}

As a result, watermarked text will statistically contain more \textit{green tokens}, an attribute unlikely to occur in human-written text.
When detecting, one can apply step (1) and (2), and calculate the z-score as the watermark strength of $\boldsymbol{x}$:
\begin{equation}
\label{eq:z-score-kgw}
s = ( |\boldsymbol{x}|_g - \gamma|\boldsymbol{x}|) / \sqrt{|\boldsymbol{x}|\gamma (1-\gamma)},
\end{equation}
where $|\boldsymbol{x}|_g$ is the number of green tokens in $\boldsymbol{x}$ and $\gamma = \frac{|\mathcal{V}_g|}{|\mathcal{V}|}$.
The presence of the watermark can be determined by comparing $s$ with a threshold.

\textbf{Unbiased watermark} (\textbf{UW}) views the process of adjusting the logits as applying a $\Delta$ function: $\boldsymbol{\tilde{z}}^{n+1} = \boldsymbol{z}^{n+1} + \Delta$, and designs a $\Delta$ function that satisfies:
\begin{equation}
    \mathbb{E}\left[\tilde{P}_M\right] = P_M,
\end{equation}
where $\tilde{P}_M$ is the probability distribution of the next token after logits adjustment~\cite{hu2023unbiased}.

\textbf{Semantic invariant robust watermark} (\textbf{SIR}) shows the robustness under re-translation and paraphrasing attack~\cite{liu2023semantic}.
Its core idea is to assign similar $\Delta$ for semantically similar prefixes.
Given prefix sequences $\boldsymbol{x}$ and $\boldsymbol{y}$, SIR adopts an embedding model $E$ to characterize their semantic similarity and trains a watermark model that yields $\Delta$ with the main objective:
\begin{equation}
    \label{eq:sir}
    \mathcal{L} = |\text{Sim}(E(\boldsymbol{x}), E(\boldsymbol{y})) - \text{Sim}(\Delta(\boldsymbol{x}), \Delta(\boldsymbol{y}))|,
\end{equation}
where $\text{Sim}(\cdot,\cdot)$ denotes similarity function.
Furthermore, $\forall i \in \{1,2,\dots,|\mathcal{V}|\}$, $\Delta{_i}$ is trained to be close to $+1$ or $-1$.
Therefore, SIR can be seen as an improvement based on KGW, where $\Delta{_i} > 0$ indicating that $v_i$ is a green token.
The original implementation of SIR uses C-BERT~\cite{chanchani-huang-2023-composition} as the embedding model, which is English-only.
To adopt SIR in the cross-lingual scenario, we use a multilingual S-BERT~\cite{reimers-gurevych-2019-sentence}\footnote{paraphrase-multilingual-mpnet-base-v2}instead.
\section{Cross-lingual Consistency of Text Watermark}
\label{sec:cross-lingual-consistency-of-text-watermark}
In this section, we define the concept of cross-lingual consistency in text watermarking and answer three research questions (RQ):
\begin{itemize}[topsep=0pt, partopsep=0pt,itemsep=0pt,parsep=4pt,leftmargin=9pt]
\item \textbf{RQ1}: To what extent are current watermarking algorithms consistent across different languages?
\item \textbf{RQ2}: Do watermarks exhibit better consistency between similar languages than between distant languages?
\item \textbf{RQ3}: Does semantic invariant watermark (SIR) exhibit better cross-lingual consistency than others (KGW and UW)?
\end{itemize}

\subsection{Definition}
We define cross-lingual consistency as the ability of a text watermark to retain its strength after the text is translated into another language.
We represent the original strength of the watermark as a random variable, denoted by $S$ (\cref{sec:details-of-watermarking-methods}), and its strength after translation as $\hat{S}$.
To quantitatively assess this consistency, we employ the following two metrics.

\paragraph{Pearson Correlation Coefficient (PCC)} We use PCC to assess linear correlation between $S$ and $\hat{S}$:
\begin{equation}
    \text{PCC}(S, \hat{S}) = \frac{\text{cov}(S, \hat{S})}{\sigma_{S}\sigma_{\hat{S}}},
\end{equation}
where $\text{cov}(S, \hat{S})$ is the covariance and $\sigma_{S}$ and $\sigma_{\hat{S}}$ are the standard deviations.
A PCC value close to $1$ suggests consistent trends in watermark strengths across languages.

\paragraph{Relative Error (RE)} Unlike PCC, which captures consistency in trends, RE is used to assess the magnitude of deviation between $S$ and $\hat{S}$:
\begin{equation}
\text{RE}(S, \hat{S}) = \mathbb{E}\left[\frac{|\hat{S} - S|}{|S|}\right] \times 100\%.
\end{equation}

A lower RE indicates that the watermark retains strength close to its original value after translation, signifying great cross-lingual consistency.

\subsection{Experimental Setup}
\label{sec:experimental-setup}
\paragraph{Setup} We sampled a subset of 500 English prompts from the mc4 dataset~\cite{2019t5}\footnote{\url{https://huggingface.co/datasets/mc4}}, and generated responses from the LLM using the text watermarking methods described in \cref{sec:watermarking-for-lms}.
The default decoding method was multinomial sampling, and both the prompts and the LLM-generated responses were in English.
To evaluate the cross-lingual consistency, these watermarked responses were translated into four languages using \texttt{gpt-3.5-turbo-0613}\footnote{\url{https://platform.openai.com/docs/models}}: Chinese (Zh), Japanese (Ja), French (Fr), and German (De).
Notably, English shares greater similarities with French and German, in contrast to its significant differences from Chinese and Japanese.

\paragraph{Models} For the LLMs, we adopt:
\begin{itemize}[leftmargin=10pt]
    \item \textsc{Baichuan-7B}~\cite{baichuan7b}: an LLM trained on 1.2 trillion tokens. It offers bilingual support for both Chinese and English officially. We also found that its vocabulary covers Japanese tokens.
    \item \textsc{Llama-2-7B}~\cite{touvron2023llama2}: trained on 2 trillion tokens and only provides support for English officially. Its vocabulary also contains tokens for European languages, such as German and French. 
\end{itemize}

\begin{table*}[t!]
    \centering
    \resizebox{0.85\linewidth}{!}{
    \begin{tabular}{l ccccc ccccc}
    \toprule
    \multirow{2}{*}{\bf Method} & \multicolumn{5}{c}{\bf PCC $\uparrow$} & \multicolumn{5}{c}{\bf RE (\%) $\downarrow$}\\
    \cmidrule(lr){2-6} \cmidrule(lr){7-11}
    &\bf En$\rightarrow$Zh &\bf  En$\rightarrow$Ja &\bf  En$\rightarrow$Fr &\bf  En$\rightarrow$De &\bf  Avg. &\bf  En$\rightarrow$Zh &\bf  En$\rightarrow$Ja &\bf  En$\rightarrow$Fr &\bf  En$\rightarrow$De &\bf  Avg.  \\
    \midrule
    \rowcolor{gray!25}
    \multicolumn{11}{c}{{\textsc{Baichuan-7B}}}\\
    KGW           &     0.074 &     0.039 &     0.034 &     0.056 &     0.051 &      87.30&\bf 91.51 &    97.78  & \bf95.35&     92.99\\
    UW            &     0.178 &     0.139 &     0.135 &     0.099 &     0.138 &      96.92&    96.62 &    98.32  &    98.60&     97.62\\
    SIR           & \bf 0.371 & \bf 0.294 &  \bf0.244 &  \bf0.226 & \bf 0.284 &  \bf 75.11&    95.36 &\bf 91.83  &    99.29& \bf 90.40\\
    \midrule
    \rowcolor{gray!25}
    \multicolumn{11}{c}{{\textsc{Llama-2-7B}}}\\
    KGW           &     0.113 &     0.108 &     0.117 &     0.065 &     0.101 & \bf 85.08 &    88.15 &    91.41  &    95.45&     90.02\\
    UW            &     0.206 &     0.206 & \bf 0.210 & \bf 0.139 &     0.190 &    102.08 &    94.35 &    94.38  &    97.83&     97.16\\
    SIR           & \bf 0.267 & \bf 0.259 &     0.186 &     0.137 & \bf 0.212 &     91.51 &\bf 73.56 &\bf 91.39  &\bf 78.79& \bf 83.81\\
    \bottomrule
    \end{tabular}}
    \caption{Comparison of cross-lingual consistency between different text watermarking methods (KGW, UW, and SIR). \textbf{Bold} entries denote the best result among the three methods.}
    \label{tab:cross-lingual-consistency}
\end{table*}

\begin{figure*}[t!]
    \centering
    \includegraphics[width=0.88\linewidth]{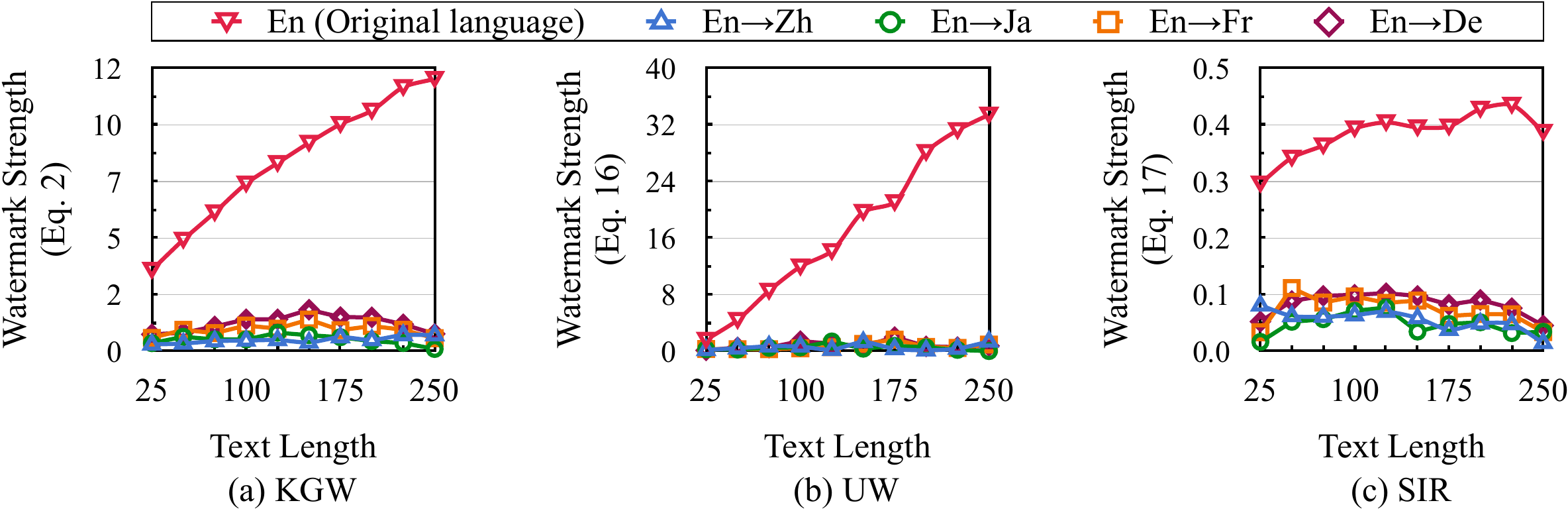}
    \caption{Trends of watermark strengths with text length before and after translation. These are the average results of \textsc{Baichuan-7B} and \textsc{Llama-2-7B}. Given the distinct calculations for watermark strengths of the three methods, the y-axis scales vary accordingly.}
    \label{fig:cross-lingual-consistency-length-overall}
\end{figure*}

\subsection{Results}
\label{sec:cross-lingual-consistency-results}
\tablename~\ref{tab:cross-lingual-consistency} presents the main results.
We forced the model to generate 200 tokens in response following the setting of~\citet{pmlr-v202-kirchenbauer23a}.
With response length restriction lifted, \figurename~\ref{fig:cross-lingual-consistency-length-overall} illustrates the trend of watermark strengths with text length.

\paragraph{Results for RQ1} We reveal a notable deficiency in the cross-lingual consistency of current watermarking methods.
Among all the settings, the PCCs are generally less than 0.3, and the REs are predominantly above 80\%.
Furthermore, \figurename~\ref{fig:cross-lingual-consistency-length-overall} visually demonstrates that the watermark strengths of the three methods exhibit a significant decrease after translation.
These results suggest that current watermarking algorithms struggle to maintain effectiveness across language translations.

\paragraph{Results for RQ2}
None of the three watermarking methods exhibits such a characteristic that its cross-lingual consistency between similar languages is significantly better than distant ones.
This means that even if two languages have similar structures or shared words, it is still difficult for watermarks to transfer between them, which poses a big challenge to the cross-lingual consistency of watermarking.

\paragraph{Results for RQ3} Overall, SIR indeed exhibits superior cross-lingual consistency compared to KGW and UW.
It achieves the best average results across the two models and two metrics.
When using \textsc{Baichuan-7B}, SIR notably outperforms other methods in terms of PCCs for all target languages.
This finding highlights the importance of semantic invariance in preserving watermark strength across languages, which we will explore more in \cref{sec:improving-cross-lingual-consistency}.
Despite its superiority, SIR still presents a notable reduction in watermark strength in cross-lingual scenarios, as evidenced by \figurename~\ref{fig:cross-lingual-consistency-length-overall}c.
\begin{figure*}[ht!]
    \centering
    \includegraphics[width=0.93\linewidth]{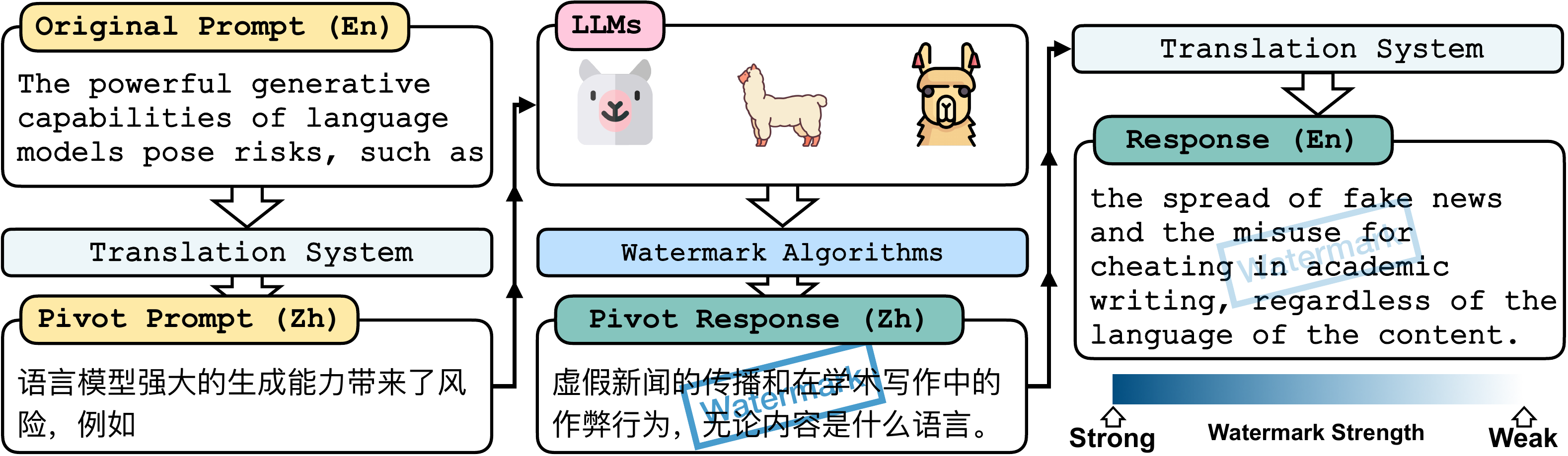}
    \caption{An example pipeline of CWRA with English (En) as the original language and Chinese (Zh) as the pivot language. When performing CWRA, the attacker not only wants to remove the watermark, but also gets a response in the original language with high quality. Its core idea is to wrap the query to the LLM into the pivot language.}
    \label{fig:cwra}
\end{figure*}

\section{Cross-lingual Watermark Removal Attack}
\label{sec:cross-lingual-watermark-removal-attack}
In the previous section, we focus on scenarios where the response of LLM is translated into other languages.
However, an attacker typically expects a response from the LLM in the same language as the prompt while removing watermarks.
To bridge this gap, we introduce the Cross-lingual Watermark Removal Attack (CWRA) in this section, constituting a complete attack process and posing a more significant challenge to text watermarking than paraphrasing and re-translation attacks.

\figurename~\ref{fig:cwra} shows the process of CWRA.
Instead of feeding the original prompt into the LLM, the attacker initiates the attack by translating the prompt into a pivot language named the pivot prompt.
The LLM receives the pivot prompt and provides a watermarked response in the pivot language.
The attacker then translates the pivot response back into the original language.
This approach allows the attacker to obtain the response in the original language.
Due to the inherent challenges in maintaining cross-lingual consistency, the watermark would be effectively eliminated during the second translation step.

\subsection{Setup}
\label{sec:setup}
To assess the practicality of attack methods, we consider two downstream tasks: text summarization and question answering.
We adopt Multi-News~\cite{fabbri-etal-2019-multi} and ELI5~\cite{fan-etal-2019-eli5} as test sets, respectively.
Both datasets are in English and require long text output with an average output length of 198 tokens.
We selected 500 samples for each test set that do not exceed the maximum context length of the model and performed zero-shot prompting on \textsc{Baichuan-7B}.
For CWRA, we select Chinese as the pivot language and compare the following two methods:
\begin{itemize}[topsep=0pt, partopsep=0pt,itemsep=0pt,parsep=4pt,leftmargin=10pt]
    \item \textbf{Paraphrase}: rephrasing the response into different wording while retaining the same meaning.
    \item \textbf{Re-translation}: translating the response into the pivot language and back to the original language.
\end{itemize}
The paraphraser and translator used in all attack methods are \texttt{gpt-3.5-turbo-0613} to ensure consistency across the different attack methods.

\subsection{Results}
\begin{figure*}[ht]
    \centering
    \includegraphics[width=0.9\linewidth]{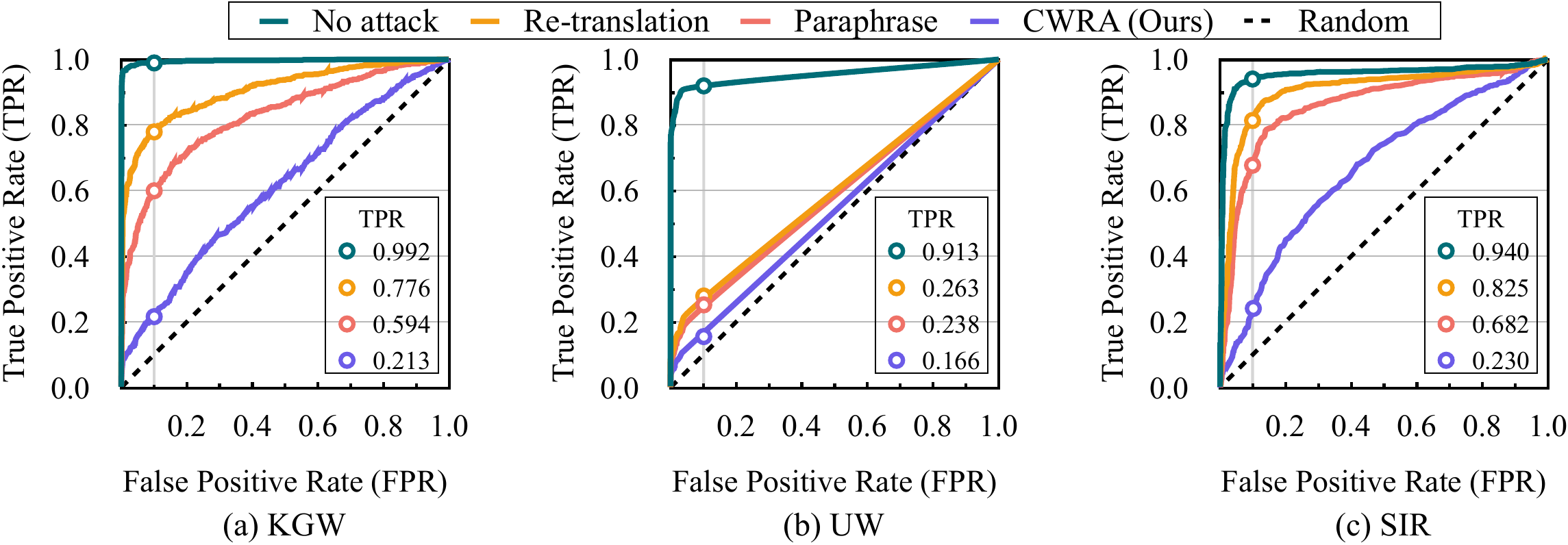}
    \caption{ROC curves for KGW, UW, and SIR under various attack methods: Re-translation, Paraphrase and CWRA. We also present TPR values at a fixed FPR of 0.1. This is the overall result of text summarization and question answering. \figurename~\ref{fig:cwra-roc-ts} and \figurename~\ref{fig:cwra-roc-qa} display results for each task.}
    \label{fig:cwra-roc}
\end{figure*}

\figurename~\ref{fig:cwra-roc} exhibits ROC curves of three watermarking methods under different attack methods.

\paragraph{CWRA vs Other Attack Methods} CWRA demonstrates the most effective attack performance, significantly diminishing the AUCs and the TPRs.
For one thing, existing watermarking techniques are not designed for cross-lingual contexts, leading to weak cross-lingual consistency.
For another thing, strategies such as Re-translation and Paraphrase are essentially semantic-preserving text rewriting.
Such strategies tend to preserve some n-grams from the original response, which may still be identifiable by the watermark detection algorithm.
In contrast, CWRA reduces such n-grams due to language switching.

\paragraph{SIR vs Other Watermarking Methods} Under the CWRA, SIR exhibits superior robustness compared to other watermarking methods.
The AUCs for KGW and UW under CWRA plummet to 0.61 and 0.54, respectively, approaching the level of random guessing.
In stark contrast, the AUC for the SIR method stands significantly higher at 0.67, aligning with our earlier observations regarding cross-lingual consistency in the RQ3 of \cref{sec:cross-lingual-consistency-results}.

\begin{table*}[ht]
    \centering
    \resizebox{0.9\linewidth}{!}{
    \begin{tabular}{l ccc  ccc  ccc}
    \toprule
    \multirow{2}{*}{\diagbox{\textbf{Attack}}{\textbf{WM}}} & \multicolumn{3}{c}{\bf KGW} & \multicolumn{3}{c}{\bf UW} & \multicolumn{3}{c}{\bf SIR} \\
    \cmidrule(lr){2-4} \cmidrule(lr){5-7} \cmidrule(lr){8-10}
    & \small\bf{\textsc{Rouge-1}} & \small\bf{\textsc{Rouge-2}} & \small\bf{\textsc{Rouge-L}} & \small\bf{\textsc{Rouge-1}} & \small\bf{\textsc{Rouge-2}} & \small\bf{\textsc{Rouge-L}} & \small\bf{\textsc{Rouge-1}} & \small\bf{\textsc{Rouge-2}} & \small\bf{\textsc{Rouge-L}} \\
    \midrule
    \rowcolor{gray!25}
    \multicolumn{10}{c}{\bf \textit{Text Summarization}} \\
    No attack      &    14.24 &     2.68 &    12.99 &    13.65 &     1.68 &    12.38 &    13.34 &    1.79 &    12.43 \\
    Re-translation &    14.11 &     2.43 &    12.89 &    13.89 &     1.77 &    12.63 &    13.63 &    1.98 &    12.61 \\
    Paraphrase     &    15.10 &     2.49 &    13.69 &    14.72 &     1.95 &    13.31 &    15.56 &    2.11 &    14.14 \\
    CWRA (Ours)    & \bf18.98 &  \bf3.63 & \bf17.33 & \bf15.88 &  \bf2.31 & \bf14.25 & \bf17.38 & \bf2.67 & \bf15.79 \\
    \midrule
    \rowcolor{gray!25}
    \multicolumn{10}{c}{\bf \textit{Question Answering}} \\
    No attack      & \bf19.00 &    2.18 &    16.09 &    11.70 &    0.49 &    9.57  &    16.95 &    1.35 &    14.91 \\
    Re-translation &    18.62 &    2.32 &    16.39 &    12.98 &    1.30 &    11.16 &    16.90 &    1.80 &    15.12 \\
    Paraphrase     &    18.45 &    2.24 & \bf16.47 &    14.38 &    1.37 &    13.07 &    17.17 &    1.79 & \bf15.54 \\
    CWRA (Ours)    &    18.23 & \bf2.56 &    16.27 & \bf15.20 & \bf1.88 & \bf13.45 & \bf17.47 & \bf2.22 &    15.53 \\

    \bottomrule
    \end{tabular}}
    \caption{Comparative analysis of text quality impacted by different watermark removal attacks.}
    \label{tab:cwra-text-quality}
\end{table*}

\paragraph{Text Quality} As shown in \tablename~\ref{tab:cwra-text-quality}, these attack methods not only preserve text quality, but also bring slight improvements in most cases due to the good translator and paraphraser.
Among the compared methods, CWRA stands out for its superior performance.
Considering that the same translator and paraphraser were used across all methods, we speculate that this is because the \textsc{Baichuan-7B} model used in our experiments performs even better in the pivot language (Chinese) than in the original language (English).
This finding implies that a potential attacker could strategically choose a pivot language at which the LLM excels to perform CWRA, thereby achieving the best text quality while removing the watermark.
\section{Improving Cross-lingual Consistency}
\label{sec:improving-cross-lingual-consistency}
Up to this point, we have observed the challenges associated with text watermarking in cross-lingual scenarios.
In this section, we first analyze two key factors essential for achieving cross-lingual consistency.
Based on our analysis, we propose a defense method against CWRA.

\subsection{Two Key Factors of Cross-lingual Consistency}
\label{sec:two-keys-factors-of-cross-lingual-consistency}
\begin{figure}[htpb]
    \centering
    \includegraphics[width=1\linewidth]{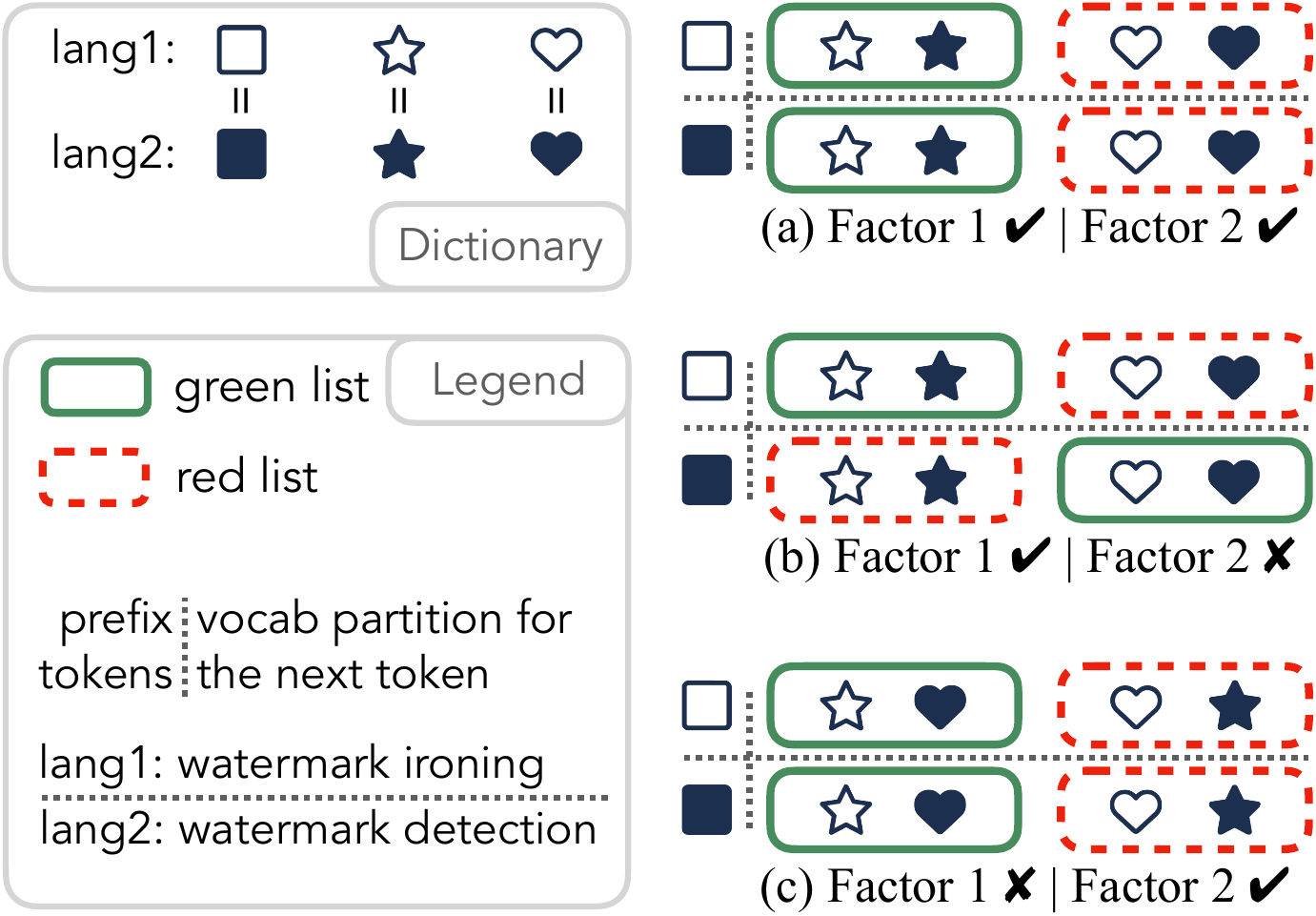}
    \caption{Three cases of attempts to maintain cross-lingual consistency. Cross-lingual consistency can only be achieved if {\solidStar} is in the green list when watermark detection. \textbf{Factor 1}: semantically equivalent tokens should be in the same list (either red or green). In these cases, {\hollowStar} = {\solidStar}, and {\hollowHeart} = {\solidHeart}. \textbf{Factor 2}: the vocabulary partitions for semantically equivalent prefixes ({\hollowSquare} and {\solidSquare}) should be the same.}
    \label{fig:defense-intro}
\end{figure}

KGW-based watermarking methods fundamentally depend on the partition of the vocabulary, i.e., the red and green lists, as discussed in \cref{sec:watermarking-for-lms}.
Therefore, cross-lingual consistency aims to achieve the following goal:
\begin{tcolorbox}[colback=lightgreen, boxrule=0pt, arc=10pt, outer arc=10pt]
\textit{The green tokens in the watermarked text will still be recognized as green tokens after being translated into other languages.}
\end{tcolorbox}
With this goal in mind, we start our analysis with a toy case in \figurename~\ref{fig:defense-intro}:
\begin{enumerate}[topsep=0pt, partopsep=0pt,itemsep=0pt,parsep=0pt,leftmargin=15pt]
    \item We define two simple languages. Language 1 (Lang1) consists of hollow tokens: {\hollowSquare}, {\hollowStar} and {\hollowHeart}. Language 2 (Lang2) consists of solid tokens: {\solidSquare}, {\solidStar} and {\solidHeart}. Tokens with the same shape are semantically equivalent.
    \item Given {\hollowSquare} as the prefix, a watermarked LM selects {\hollowStar} from [{\hollowStar}, {\solidStar}, {\hollowHeart}, {\solidHeart}]\footnote{We suppose that this LM is bilingual and omit the prefix tokens {\hollowSquare} and {\solidSquare} for simplicity.} as the next token. Due to watermarking, {\hollowStar} is a green token.
    \item A machine translator (MT) then translates the entire sentence ``{\hollowSquare} {\hollowStar}'' into Lang2: ``{\solidSquare} {\solidStar}''.
\end{enumerate}

The question of interest is: what conditions must the vocabulary partition satisfy so that the token {\solidStar}, the semantic equivalent of {\hollowStar}, is also included in the green list?

\figurename~\ref{fig:defense-intro}(a) illustrates a successful case, where two key factors exists:
\begin{enumerate}[topsep=0pt, partopsep=0pt,itemsep=0pt,parsep=0pt,leftmargin=15pt]
    \item \textbf{Cross-lingual semantic clustering of the vocabulary}: semantically equivalent tokens must be in the same partition, either green or red lists.
    \item \textbf{Cross-lingual semantic robust vocabulary partition}: for semantically equivalent prefixes in different languages: {\hollowSquare} and {\solidSquare}, the partitions of the vocabulary must be the same.
\end{enumerate}
Both \figurename~\ref{fig:defense-intro}(b) and \figurename~\ref{fig:defense-intro}(c) satisfy only one of the two factors, thus failing to recognize {\solidStar} as a green token and losing cross-lingual consistency.

\subsection{Defense Method against CWRA}
We now improve the SIR so that it satisfies the two factors described above.
In \figurename~\ref{fig:defense-intro}, the tokens of the two languages are equivalent in one-to-one correspondence.
In practice, we relax this semantic equivalence into semantic similarity.

As discussed in \cref{sec:watermarking-for-lms}, SIR uses the $\Delta$ function to represent vocabulary partition ($\Delta\in\mathbb{R}^{|\mathcal{V}|}$), where $\Delta_i > 0$ indicating that $v_i$ is a green token.
Based on \cref{eq:sir}, SIR has already optimized for Factor 2 when using a multilingual embedding model.
For prefixes $\boldsymbol{x}$ and $\boldsymbol{y}$, the similarity of their vocabulary partitions for the next token should be close to their semantic similarity:
\begin{equation}
    \text{Sim}(\Delta(\boldsymbol{x}), \Delta(\boldsymbol{y}))
    \approx
    \text{Sim}(E(\boldsymbol{x}), E(\boldsymbol{y})),
\end{equation}
where $E$ is a multilingual embedding model.

Based on SIR, we focus on Factor 1, i.e., cross-lingual semantic clustering of the vocabulary.
Formally, we define semantic clustering as a partition $\mathcal{C}$ of the vocabulary $\mathcal{V}$:
$
    \mathcal{C}=\{\mathcal{C}_1,\mathcal{C}_2,\dots,\mathcal{C}_{|\mathcal{C}|}\}
$
, where each cluster $\mathcal{C}_k$ consists of semantically similar tokens.
Instead of assigning biases for each token in $\mathcal{V}$, we adapt the $\Delta$ function so that it yields biases to each cluster in $\mathcal{C}$, i.e., $\Delta\in \mathbb{R}^{|\mathcal{C}|}$.
Thus, the process of adjusting the logits should be:
\begin{equation}
    \begin{aligned}
        &\forall i \in \{1,2,\dots,|\mathcal{V}|\}, \\
        &\boldsymbol{\tilde{z}}^{n+1}_{i} = \boldsymbol{z}^{n+1}_{i} + \Delta_{C(i)},\\
    \end{aligned}
    \end{equation}
where $C(i)$ indicates the index of $v_i$'s cluster within $\mathcal{C}$.
By doing so, if token $v_i$ and $v_j$ are semantically similar, they will receive the same bias on logits:
\begin{equation}
    C(i) = C(j) \implies \Delta_{C(i)} = \Delta_{C(j)}.
\end{equation}
In other words, if $v_i$ and $v_j$ are translations of each other, they will fall into the same list.
Algorithm~\ref{alg:constructing-semantic-clusters} shows the procedure of semantic clustering.
To obtain such a semantic clustering $\mathcal{C}$, we treat each token in $\mathcal{V}$ as a node, and add an edge $(v_i, v_j)$ whenever $(v_i, v_j)$ corresponds to an entry in a bilingual dictionary $\mathcal{D}$.
Therefore, $\mathcal{C}$ is all the connected components of this graph.
\begin{algorithm}[H]
    \caption{Constructing semantic clusters}
    \label{alg:constructing-semantic-clusters}
    \begin{algorithmic}[1]
        \Require the vocabulary of the LM $\mathcal{V}$, a bilingual dictionary $\mathcal{D}$
        \Ensure Semantic clusters $\mathcal{C}$
        \State Initialize a graph $G=(\mathcal{V}, \varnothing)$
        \For{each entry $(v_i, v_j)$ in $\mathcal{D}$}
            \If{$v_i \in \mathcal{V}$ and $v_j \in \mathcal{V}$}
                \State Add an edge $(v_i, v_j)$ to $G$
            \EndIf
        \EndFor
        
        \State $\mathcal{C} = \varnothing$
        \For{each connected component $\mathcal{C}_k$ in $G$}
            \State $\mathcal{C} = \mathcal{C}\cup\{\mathcal{C}_k\}$
        \EndFor
        \State \Return $\mathcal{C}$
    \end{algorithmic}
\end{algorithm}
\paragraph{Setup} We name this method as X-SIR, implemented it with an unified external dictionary $\mathcal{D}$ that covers English (En), Chinese (Zh), Japanese (Ja), French (Fr), and German (De).
Following the settings of \cref{sec:experimental-setup} and \cref{sec:setup}, we apply X-SIR on \textsc{Baichuan-7B}, analyze it in the following and detail its limitations in \cref{sec:limitations}.
We also present more results on other LLMs in \cref{sec:appendix-xsir}.

\begin{figure}[htpb]
    \centering
    \includegraphics[width=\linewidth]{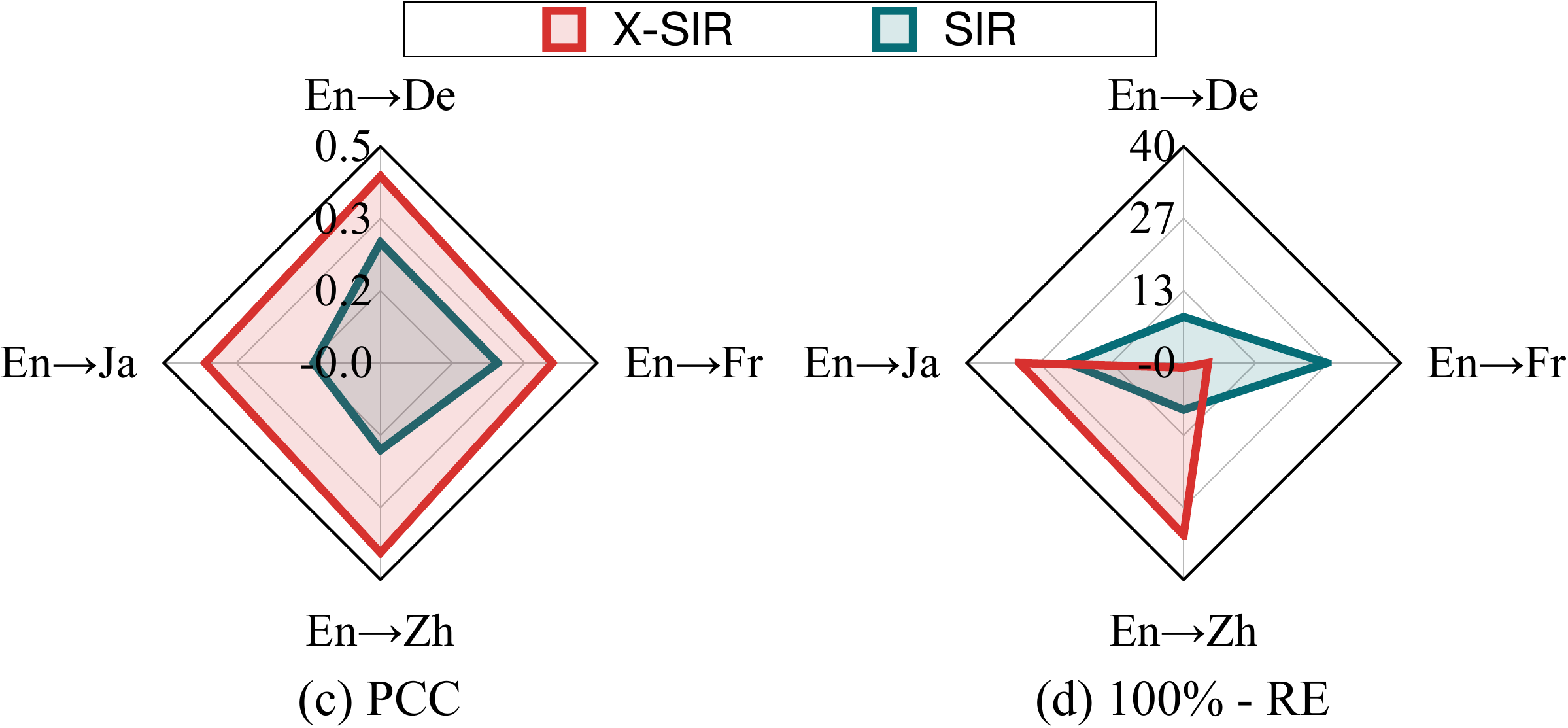}
    \caption{Cross-lingual consistency in terms of PCC and RE. We follow the same setting as \cref{sec:experimental-setup} and plot 100\%-RE for visualization.}
    \label{fig:pcc-re}
\end{figure}

\begin{figure}[htpb]
    \centering
    \begin{subfigure}[b]{\linewidth}
        \centering
        \includegraphics[width=\linewidth]{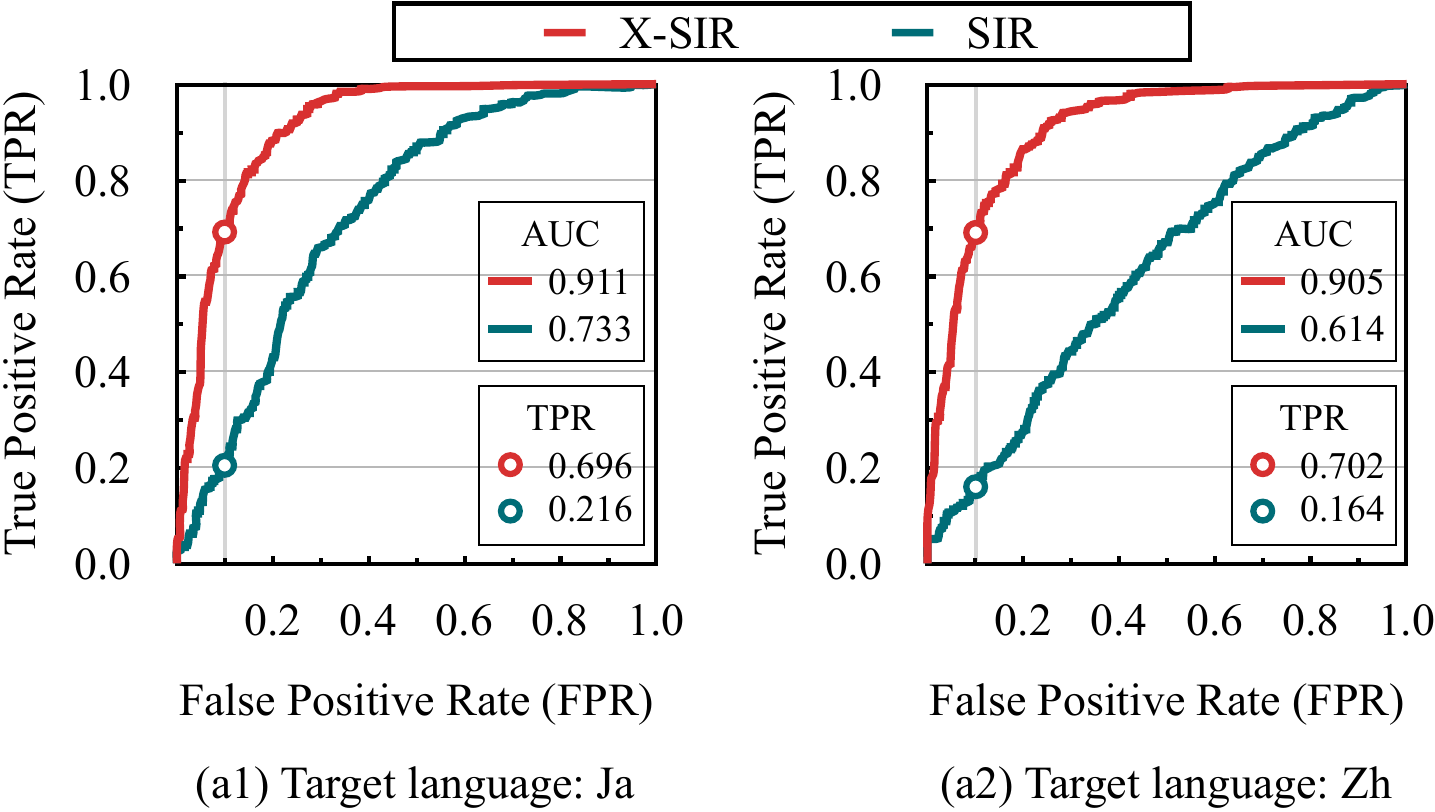}
        \caption{Direct translation}
        \label{fig:xsir-roc-direct-translation}
    \end{subfigure}

    \begin{subfigure}[b]{\linewidth}
        \centering
        \includegraphics[width=\linewidth]{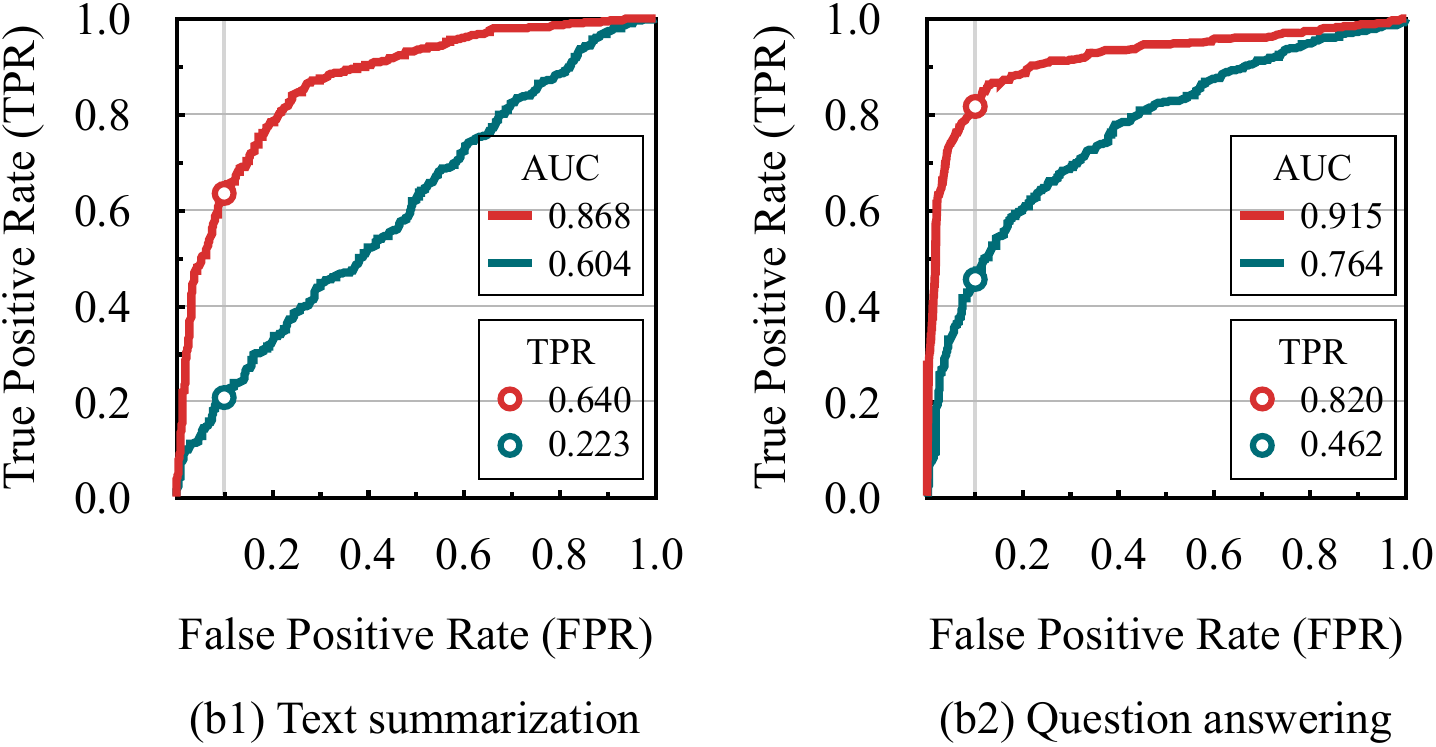}
        \caption{CWRA}
        \label{fig:xsir-roc-cwra}
    \end{subfigure}
    \caption{Watermark detection performance under cross-lingual scenarios. (a) Direct translation: the original English responses are translated into Ja or Zh following the setting in \cref{sec:experimental-setup}. (b) CWRA: the full CWRA process as we did in~\cref{sec:cross-lingual-watermark-removal-attack}, where English is the original language and Chinese is the pivot language.}
    \label{fig:xsir-watermark-detection-performance}
\end{figure}

\paragraph{Cross-lingual consistency \& ROC curves}
\figurename~\ref{fig:pcc-re} shows the cross-lingual consistency of SIR and X-SIR when the original responses are translated into other languages.
Considering both PCC and RE, X-SIR notably improves the cross-lingual consistency over SIR when the target language is Zh or Ja.
However, for De and Fr, X-SIR improves only the PCC but not RE, which we regard as a limitation and discuss in~\cref{sec:limitations}.
Further, \figurename~\ref{fig:xsir-watermark-detection-performance} evaluates both methods' watermark detection performance under cross-lingual scenarios.
Whether directly translating the responses into other languages (\figurename~\ref{fig:xsir-roc-direct-translation}) or using CWRA (\figurename~\ref{fig:xsir-roc-cwra}), X-SIR substantially enhances the watermark detection performance, with an average increase in TPR by 0.448 and AUC by 0.221.
These findings validate the two key factors of cross-lingual consistency that we identified in~\cref{sec:two-keys-factors-of-cross-lingual-consistency}.

\begin{figure}[htpb]
    \centering
    \includegraphics[width=\linewidth]{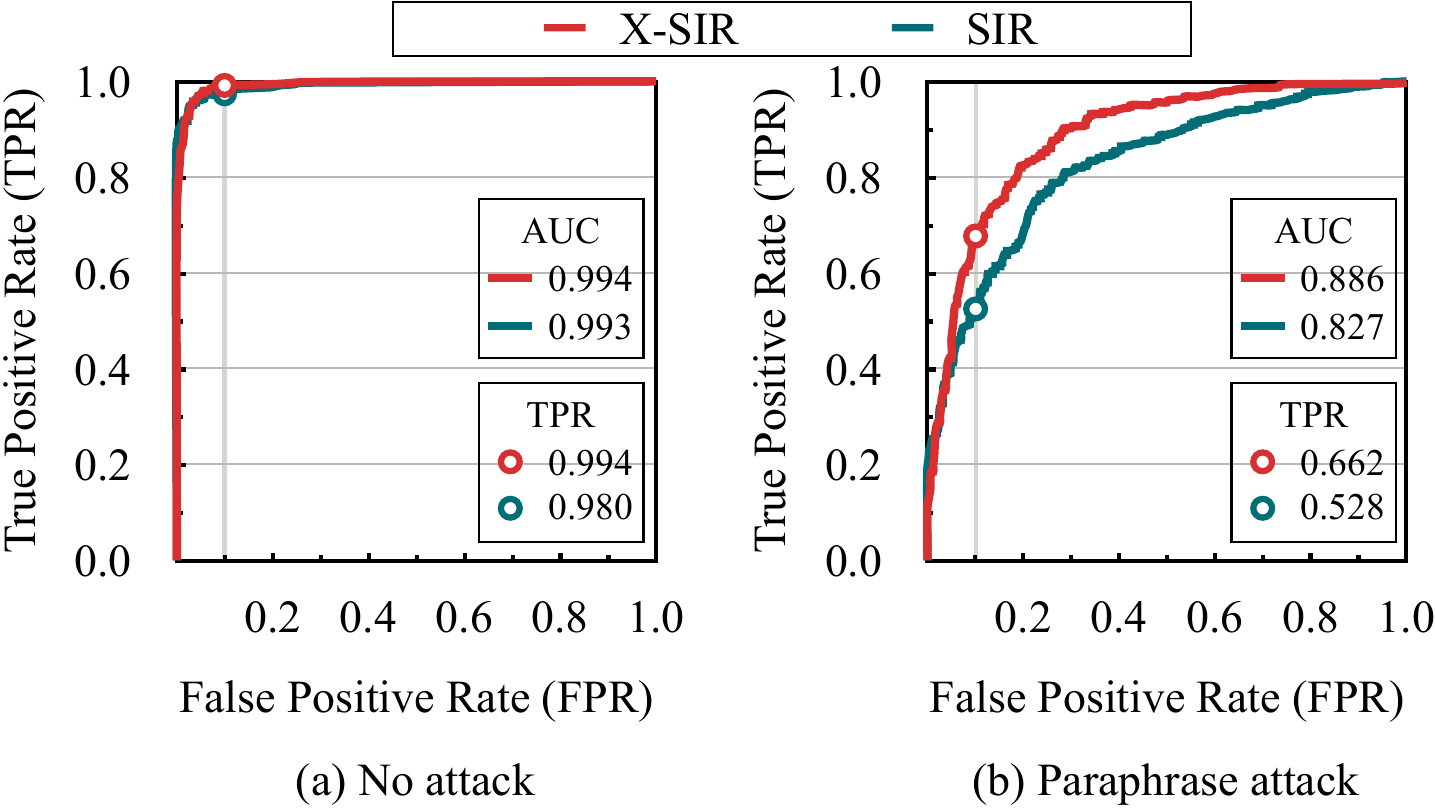}
    \caption{Watermark detection performance under no attack and paraphrase attack.}
    \label{fig:xsir-no-and-para-attack}
\end{figure}

\paragraph{No attack \& Paraphrase attack}
The next question is whether the addition of semantic clustering of the vocabulary to X-SIR will affect the original detection performance of SIR.
As shown in \figurename~\ref{fig:xsir-no-and-para-attack}, when there is no attack, both methods perform comparably.
However, under the paraphrase attack, X-SIR performs slightly better than SIR.
This is reasonable because paraphrasing can be regarded as a special kind of ``translation'' within the same language and X-SIR improves such intra-lingual consistency.

\paragraph{Text quality}
As shown in \tablename~\ref{tab:x-sir-text-quality}, X-SIR achieves better text quality than SIR in text summarization, and comparable performance on question answering, meaning the semantic clustering of vocabulary will not negatively affect text quality.
\begin{table}[htpb]
    \centering
    \resizebox{\linewidth}{!}{
    \begin{tabular}{r c c c}
    \toprule
    \bf Method & \bf{\textsc{Rouge-1}} & \bf{\textsc{Rouge-2}} & \bf{\textsc{Rouge-L}}\\
    \midrule
    \rowcolor{gray!25}
    \multicolumn{4}{c}{\bf \textit{Text Summarization}} \\
    SIR   &     13.34 &    1.79 &    12.43  \\
    X-SIR &  \bf15.65 & \bf2.04 & \bf14.29  \\
    \midrule
    \rowcolor{gray!25}
    \multicolumn{4}{c}{\bf \textit{Question Answering}} \\
    SIR   & \bf16.95 &    1.35 & \bf14.91  \\
    X-SIR &    16.77 & \bf1.39 &    14.07  \\
    \bottomrule
    \end{tabular}}
    \caption{Effects of X-SIR and SIR on text quality.}
    \label{tab:x-sir-text-quality}
\end{table}

\section{Related Work}
\subsection{LLM Watermarking}
\label{sec:llm-watermarking}
Text watermarking aims to embed a watermark into a text and detect the watermark for any given text.
Currently, text watermark method can be classified into two categories~\cite{liu2023survey}: watermarking for existing text and watermarking for generated text.
In this work, we focus on the latter, which is more challenging and has more practical applications.

This type of watermark method usually can be illustrated as the watermark ironing process (modifying the logits of the LLM during text generation) and watermark detection process (assess the presence of watermark by a calculated watermark strength score).
\citet{pmlr-v202-kirchenbauer23a} introduces KGW, the first watermarking method for LLMs.
\citet{hu2023unbiased} proposes UW without affecting the output probability distribution compared to KGW.
\citet{liu2023semantic} introduces SIR, a watermarking method taking into account the semantic information of the text, which shows robustness to text re-writing attacks.
\citet{liu2023unforgeable} proposes the first unforgeable and publicly verifiable watermarking algorithm for LLMs.
SemStamp~\cite{hou2023semstamp} is another semantic-related watermarking method and it generate watermarked text at sentence granularity instead of token granularity.
\citet{tu2023waterbench} introduces WaterBench, the first comprehensive benchmark for LLM watermarks.
\subsection{Watermark Robustness}
\label{sec:watermark-robustness}
A good watermarking method should be robust to various watermarking removal attacks.
However, current works on watermarking robustness mainly focus on single-language attacks, such as paraphrase attacks.
For example, \citet{kirchenbauer2023reliability} evaluates the robustness of KGW against paraphrase attacks as well as copy-paste attacks and proposes a detect trick to improve the robustness to copy-paste attacks.
\citet{zhao2023provable} employs a fixed green list to improve the robustness of KGW against paraphrase attacks and editing attacks.
\citet{chen2023xmark} proposes a new paraphrase robust watermarking method ``XMark'' based on ``text redundancy'' of text watermark.
\citet{lu2024entropy} proposes an entropy-based text watermarking detection method that achieves better detection performance in low-entropy scenarios.
\section{Conclusion}
This work aims to investigate the cross-lingual consistency of watermarking methods for LLMs.
We first characterize and evaluate the cross-lingual consistency of current watermarking techniques, revealing that current watermarking methods struggle to maintain their watermark strengths across different languages.
Based on this observation, we propose the cross-lingual watermark removal attack (CWRA), which significantly challenges watermark robustness by efficiently eliminating watermarks without compromising text quality.
Through the analysis of two primary factors that influence cross-lingual consistency, we propose X-SIR as a defense strategy against CWRA.
Despite its limitations, this approach greatly improves watermark detection performance under cross-lingual scenario and paves the way for future research.
Overall, this work completes a closed loop in the study of cross-lingual consistency in watermarking, including: evaluation, attacking, analysis, and defensing.

Since our first release, the X-SIR algorithm has been integrated into MarkLLM~\cite{pan2024markllm}, an open-source toolkit for LLM Watermarking.
\section{Limitations}
\label{sec:limitations}
X-SIR relies on the semantic clustering described in Algorithm~\ref{alg:constructing-semantic-clusters} which only considers tokens shared by the vocabulary $\mathcal{V}$ of the model and the external dictionary $\mathcal{D}$.
This design results in the following limitations:
\begin{itemize}[topsep=0pt, partopsep=0pt,itemsep=0pt,parsep=4pt,leftmargin=10pt]
    \item \textbf{Language coverage}: X-SIR only supports languages supported by the model. However, in a real scenario, an attacker can choose the original and the pivot language at will. This limitation has revealed by \figurename~\ref{fig:pcc-re}, where X-SIR brings a much more significant boost to Ja \& Zh than to De \& Fr since the vocabulary of the LLM (\textsc{Baichuan-7B}) supports Zh \& Ja better.
    \item \textbf{Vocab coverage}: Since the external dictionary $\mathcal{D}$ only contains whole words, word units can not be clustered in Algorithm~\ref{alg:constructing-semantic-clusters} and will left as isolated nodes. Consequently, X-SIR's performance might be compromised if the tokenizer favors finer-grained token segmentation.
\end{itemize}
X-SIR does not solve the issue of cross-lingual consistency but sets the stage for future research.

\section*{Acknowledgements}
This paper is mainly supported by the Tencent AI Lab Fund (RBFR2024002). Zhiwei and Rui are partially supported by the National Natural Science Foundation of China (62176153) and the Shanghai Municipal Science and Technology Major Project (2021SHZDZX0102, as the MoE Key Lab of Artificial Intelligence, AI Institute, Shanghai Jiao Tong University). Zhuosheng is partially supported by the Joint Funds of the National Natural Science Foundation of China (Grant No. U21B2020).

\bibliography{anthology,custom}

\clearpage
\appendix
\section{Details of Watermarking Methods}
\label{sec:details-of-watermarking-methods}
As discussed in \cref{sec:llm-watermarking}, we focus on the watermarking methods for large language models (LLMs). 
A watermarking method can be divided into two processes: the watermark ironing process and the watermark detection process.
In ironing process, the watermark is embedded into the text by modifying the logits of the LLM during text generation.
In detection process, the watermark detector calculates the \textbf{watermark strength score} $s$ to assess the presence of watermark.
$s$ is a scalar value to indicate the strength of the watermark in the text.
For any given text, we can calculate its watermark strength score $s$ based on detection process of the watermarking method.
A higher $s$ indicates that the text is more likely to contain watermark.
In the opposite, a lower $s$ indicates that the text is less likely to contain watermark.
Every watermarking method has its own way to ironing the watermark and calculate the watermark strength score $s$.
We detail KGW, UW and SIR in the following sections.

\subsection{KGW}
\label{sec:appendix-kgw}
In \cref{sec:watermarking-for-lms}, we introduce the processes of watermark ironing and watermark detection in the KGW method. 
Here, we detail the experimental settings employed for KGW.
KGW uses a hash function $H$ to compute the hash of the previous $k$ tokens. 
In this work, we adhere to the experimental setting reported by~\citet{pmlr-v202-kirchenbauer23a}, employing the hash function \textbf{minhash} with $k = 4$. 
The ratio of green token lists $\mathcal{V}_g$ to the total word list $\mathcal{V}$ is set at $\gamma = 0.25$. 
Additionally, the constant bias $\delta$ is fixed at $2.0$.
\subsection{UW}
\label{sec:appendix-uw}
The ironing process in UW is analogous to that in KGW; however, the two differ in their respective function of modifying the logits. Here is the detail watermark ironing process in UW:
\begin{enumerate}[topsep=0pt, partopsep=0pt,itemsep=0pt,parsep=0pt]
    \item[(1)] compute a hash of $\boldsymbol{x}^{1:n}$: $h^{n+1}=H(\boldsymbol{x}^{1:n})$, and use $h^{n+1}$ as seed generating a random number $p\in\left[0,1\right)$.
    \item[(2)] determine the token $t$ satisfies:
        \begin{align}
        p\in &\left[\sum_{i=1}^{t-1}P_{Mi}(x^{n+1} | \boldsymbol{x}^{1:n}),\right.\nonumber\\
        &\left. \sum_{i=1}^{t}P_{Mi}(x^{n+1} | \boldsymbol{x}^{1:n})\right)
        \end{align}
    \item[(3)] set $P_{Mi}(x^{n+1} | \boldsymbol{x}^{1:n})=0$ for $i\neq t$ and $P_{Mt}(x^{n+1} | \boldsymbol{x}^{1:n})=1$.
\end{enumerate}
Then we get the adjusted probability of next token $\tilde{P}_M(x^{n+1} | \boldsymbol{x}^{1:n})$.

The detection process calculates a maximin variant Log Likelihood Ratio (LLR) of the detected text to assess the watermark strength score.
Log Likelihood Ratio (LLR) is defined as: 
\begin{equation}
    r_i=\frac{\tilde{P}_M(x^i|\boldsymbol{x}^{1:i-1})}{P_M(x^i|\boldsymbol{x}^{1:i-1})}
\end{equation}
The total score is defined as:
\begin{equation}
    R=\frac{\tilde{P}_M(\boldsymbol{x^{a+1:n}}|\boldsymbol{x}^{1:a})}{P_M(\boldsymbol{x^{a+1:n}}|\boldsymbol{x}^{1:a})}
\end{equation}
Where $\boldsymbol{x^{1:a}}$ is prompt and $\boldsymbol{x^{a+1:n}}$ is the detected text.
Let 
\begin{align}
    P_i&=P_M(x^i|\boldsymbol{x^{1:i-1}})\\
    Q_i&=\tilde{P}_M(x^i|\boldsymbol{x^{1:i-1}})\\
    R_i&=(r_i(x_1),r_i(x_2),\cdots,r_i(x_{|\mathcal{V}|}))
\end{align}
Where $r_i(x_k)$ is the LLR of token $x_k$ at position $i$.
UW use a maximin variant LLR ~\cite{hu2023unbiased} to avoid the limitation of the origin LLR.
The calculating process of maximin variant LLR can be formulated as follows:
\begin{align}
    &\max_{R_i}\min_{Q^{'}_{i}\in\Delta_\mathcal{V}, TV(Q^{'}_{i}, Q_{i})\leq d} \left\langle Q_i^{\prime}, R_i\right\rangle, \quad \nonumber\\
    &\text { s.t. }\left\langle P_i, \exp \left(R_i\right)\right\rangle \leq 1
\end{align}
Where $\Delta_\mathcal{V}$ is the set of all probability distributions over the symbol set $\mathcal{V}$, and $TV$ is the total variation distance, $d$ is a hyperparameter to control $TV$, and $\left\langle \cdot, \cdot\right\rangle$ is the inner product.
UW utilizes the maximin variant LLR to calculate the watermark strength score.

In the experiments, we follow the experiment settings of original paper, using the previous $5$ tokens to compute the hash and set $d=0$.
\subsection{SIR}
\label{sec:appendix-sir}
As introduced in \cref{sec:watermarking-for-lms}, the ironing process in SIR assigns a watermark bias, $\Delta_{i}$, to every token $v_i$. 

For any given text, the watermark detector calculates the mean watermark bias to determine the watermark strength score. 
Consider the detected text represented as $\boldsymbol{x}=\left(x^1,\ldots,x^{N}\right)$. The watermark strength score, $s$, can be expressed by the following equation:
\begin{equation}
    s = \frac{\sum_{n=1}^{N} \Delta_{I(x^n)}(\boldsymbol{x}^{1:n-1})}{N},
\end{equation}
where $I(x^n)$ indicates the index of the token $x^n$ within the vocabulary $\mathcal{V}$, and $\Delta_{I(x^n)}(\boldsymbol{x}^{1:n-1})$ denotes the watermark bias for token $x^n$ at position $n$. 
Given that the watermark bias satisfies the unbiased property, $\sum_{t \in \mathcal{V}} \Delta_{I(t)} = 0$, the expected detection score for normal text is $0$.
Consequently, the detection score for watermarked text should exceed $0$ significantly.

\section{Full Result of Watermark Removal Attack}
\label{sec:appendix-cwra}
\begin{figure*}[t]
    \centering
    \begin{subfigure}[b]{\linewidth}
        \centering
        \includegraphics[width=\linewidth]{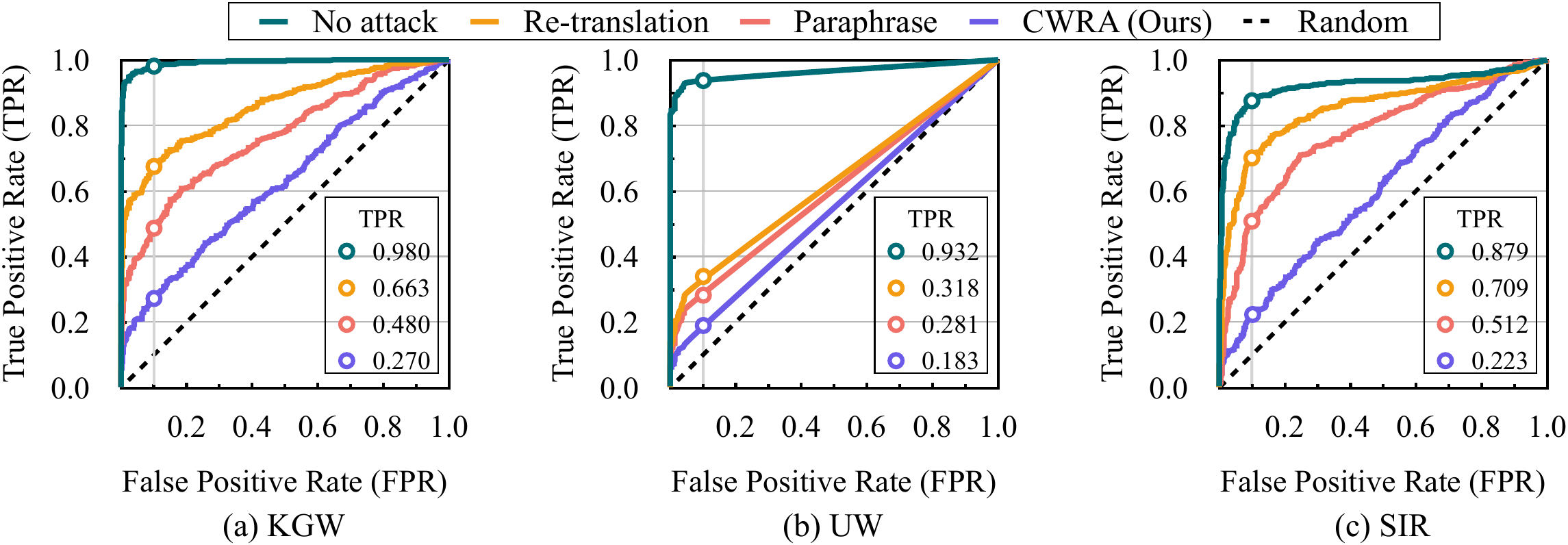}
        \caption{Text summarization}
        \label{fig:cwra-roc-ts}
    \end{subfigure}
    \begin{subfigure}[b]{\linewidth}
        \centering
        \includegraphics[width=\linewidth]{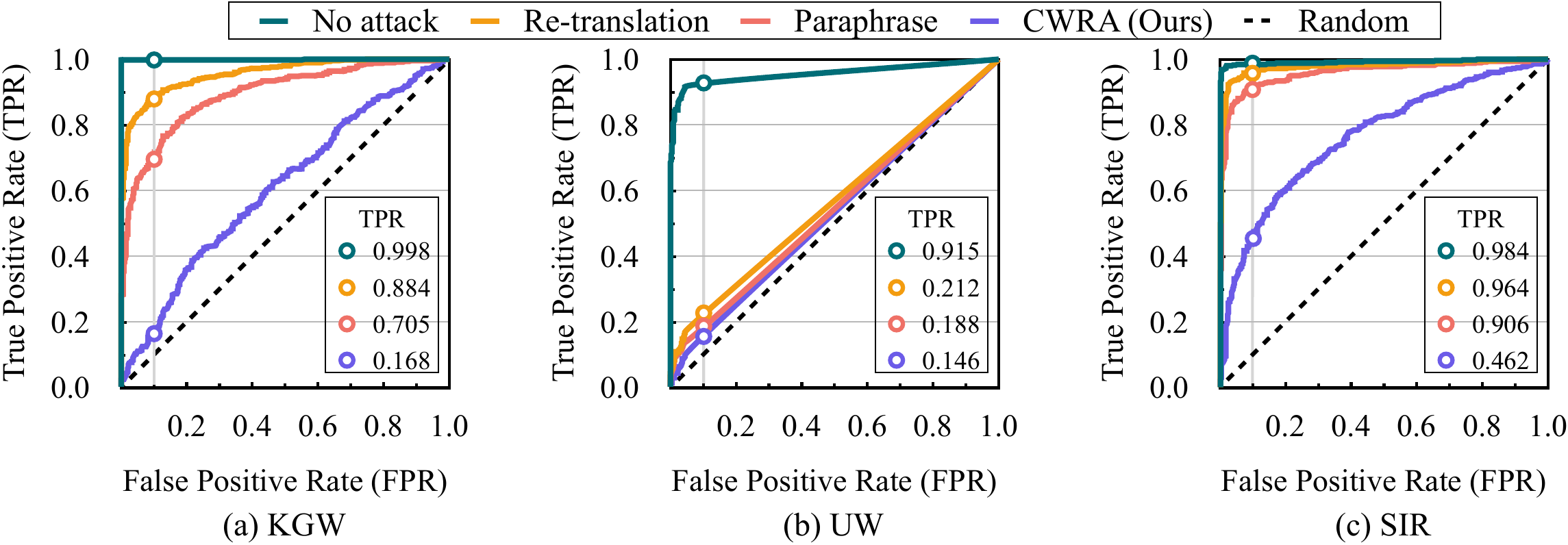}
        \caption{Question answering}
        \label{fig:cwra-roc-qa}
    \end{subfigure}
    \caption{ROC curves for KGW, UW and SIR under various attack methods: Re-translation, Paraphrase and CWRA.
    We also present TPR values at a fixed FPR of $0.1$.}
    \label{fig:cwra-roc-app}
\end{figure*}

\figurename~\ref{fig:cwra-roc-app} presents the full results of watermark removal attacks.
It is equivalent to \figurename~\ref{fig:cwra-roc}, but presents results on text summarization and question answering separately.

\section{Full Result of X-SIR}
\label{sec:appendix-xsir}
\begin{figure*}[t]
    \centering
    \begin{subfigure}[b]{\linewidth}
        \centering
        \includegraphics[width=\linewidth]{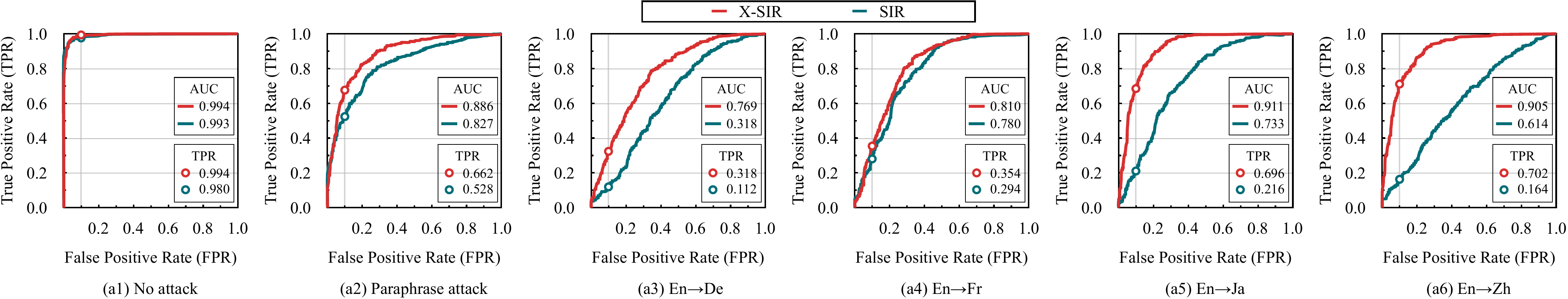}
        \caption{\textsc{Baichuan-7B}}
        \label{fig:all-roc-curves-baichuan}
    \end{subfigure}
    \vspace{1pt}

    \begin{subfigure}[b]{\linewidth}
        \centering
        \includegraphics[width=\linewidth]{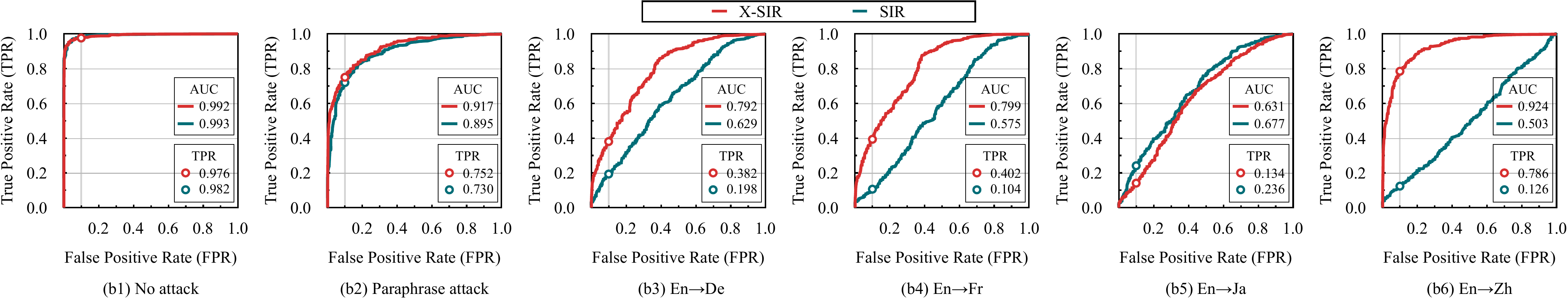}
        \caption{\textsc{Baichuan2-7B-Base}}
        \label{fig:all-roc-curves-baichuan2}
    \end{subfigure}
    \vspace{1pt}

    \begin{subfigure}[b]{\linewidth}
        \centering
        \includegraphics[width=\linewidth]{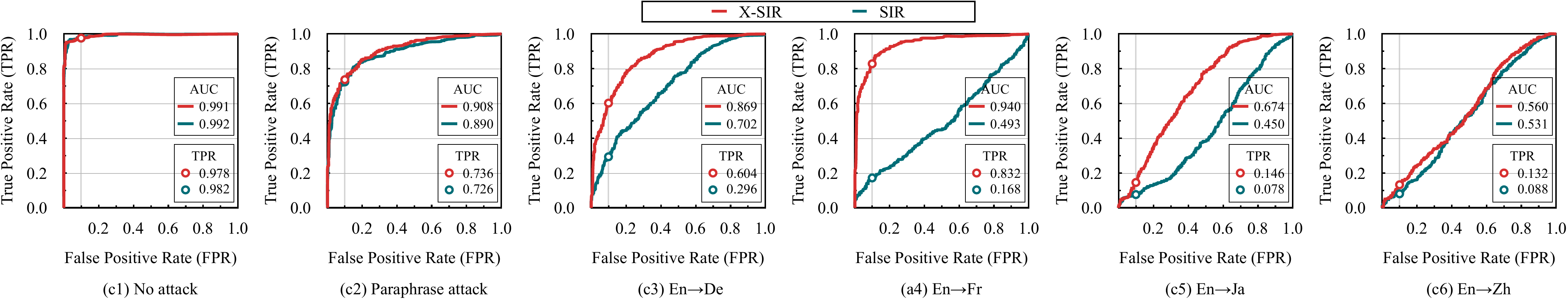}
        \caption{\textsc{Llama-2-7B}}
        \label{fig:all-roc-curves-llama2}
    \end{subfigure}
    \vspace{1pt}

    \begin{subfigure}[b]{\linewidth}
        \centering
        \includegraphics[width=\linewidth]{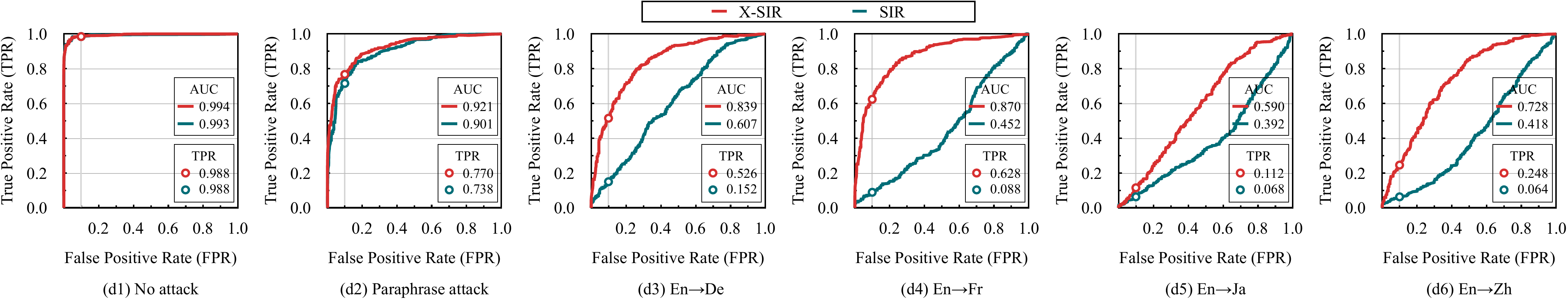}
        \caption{\textsc{Mistral-7B-v0.1}}
        \label{fig:all-roc-curves-mistral}
    \end{subfigure}
    \caption{ROC curves for X-SIR and SIR under various attack methods: no attack, paraphrase and direct translation. We also present AUC and TPR values at a fixed FPR of 0.1.}
    \label{fig:all-roc-curves}
\end{figure*}

\figurename~\ref{fig:all-roc-curves} presents full watermark detection performance (ROC curves, AUCs and TPRs) of X-SIR.
We tested it on the following LLMs:
\begin{itemize}[leftmargin=10pt]
    \item \textsc{Baichuan-7B}~\cite{baichuan7b}\footnote{\url{https://huggingface.co/baichuan-inc/Baichuan-7B}}
    \item \textsc{Baichuan2-7B-Base}~\cite{yang2023baichuan}\footnote{\url{https://huggingface.co/baichuan-inc/Baichuan2-7B-Base}}
    \item \textsc{Llama-2-7B}~\cite{touvron2023llama2}\footnote{\url{https://huggingface.co/meta-llama/Llama-2-7b-hf}}
    \item \textsc{Mistral-7B-v0.1}~\cite{jiang2023mistral}\footnote{\url{https://huggingface.co/mistralai/Mistral-7B-v0.1}}.
\end{itemize}
We tested it under the following attack methods:
\begin{itemize}[leftmargin=10pt]
    \item No attack
    \item Paraphrase attack (\cref{sec:setup})
    \item Direct translation (En$\rightarrow$XX): translating original English responses to other languages.
\end{itemize}
We also plot the performance of SIR as baselines.

\end{document}